\documentclass[11pt]{article}

\usepackage[final]{acl}
\usepackage[table]{xcolor}

\usepackage{times}
\usepackage{latexsym}
\usepackage{booktabs}
\usepackage{hyperref}
\usepackage{graphicx}
\usepackage{multirow}
\usepackage{amsmath,amssymb,amsfonts}
\usepackage{mathtools}
\usepackage{makecell}
\usepackage{bm}
\usepackage{booktabs}
\usepackage{algorithm}
\usepackage{subcaption}
\usepackage{xspace}
\usepackage{algorithmic}

\newcommand{\best}[1]{\textbf{#1}}

\usepackage[T1]{fontenc}

\usepackage[utf8]{inputenc}

\usepackage{microtype}

\usepackage{inconsolata}

\usepackage{graphicx}

\title{LiST: Local-Simplex Test-Time LoRA Fusion}

\author{
 \textbf{Yihua Shao\textsuperscript{2,3,*}},
 \textbf{Jia Li\textsuperscript{4,*}},
 \textbf{Siyu Chen\textsuperscript{3}},
 \textbf{Xinyu Luo\textsuperscript{5}},
 \textbf{Yang Liu\textsuperscript{6}},
 \textbf{Kecheng Chen\textsuperscript{5}},
  \\
 \textbf{Xinwei Long\textsuperscript{7}},
 \textbf{Lingyu Zhu\textsuperscript{5}},
 \textbf{Fanhu Zeng\textsuperscript{3}},
 \textbf{Maolin Wang\textsuperscript{5}},
 \textbf{Ziyang Yan\textsuperscript{8}},
 \\
 \textbf{Jingcai Guo\textsuperscript{2}},
 \textbf{Hao Tang\textsuperscript{9}},
 \textbf{Nicu Sebe\textsuperscript{10}},
 \textbf{Zhenyi Wang\textsuperscript{1}}
\\
\\
 \textsuperscript{1}SUSTech,
 \textsuperscript{2}PolyU,
 \textsuperscript{3}CASIA,
 \textsuperscript{4}GDUT,
 \textsuperscript{5}CityU,
 \textsuperscript{6}BigAI,
 \textsuperscript{7}THU,
 \textsuperscript{8}TAU,
 \textsuperscript{9}PKU,
 \textsuperscript{10}UniTrento
\\
\small{\textsuperscript{*}Equal contribution.}\\
 \small{
   \textbf{Correspondence:} Zhenyi Wang, \href{mailto:email@domain}{wangzy9@sustech.edu.cn} and Jingcai Guo, \href{mailto:email@domain}{jc-jingcai.guo@polyu.edu.hk}
 }
}

\begin{document}
\maketitle
\begin{abstract}
Task-specific LoRA adapters offer a modular way to specialize large language and vision-language models. However, existing adapter composition methods are mostly static and cannot adapt to individual test inputs. To address these issues, we propose \textbf{LiST}, a label-free test-time LoRA fusion framework that converts an existing LoRA bank into a target-conditioned local simplex and searches sample-specific fusion weights at inference time. LiST builds joint task representations from LoRA parameter anchors and prompt-level behavior vectors, retrieves neighboring adapters as a local search space, and performs branch-preserving fusion without updating the backbone or adapters. Candidate weights are selected by a prompt-level energy with prior, geometric, and stochastic-consistency constraints, and are deployed only when they pass a safe acceptance rule. Otherwise, LiST falls back to a target-conditioned prior. Experiments on multimodal and language benchmarks show that LiST outperforms static LoRA merging and conventional test-time adaptation baselines, while preserving task-specific adapter utility and improving robustness on unseen tasks. Code is available at \href{https://github.com/YihuaJerry/LiST}{https://github.com/YihuaJerry/LiST}
\end{abstract}

\section{Introduction}

Multimodal Large Language Model~(MLLMs)~\cite{liu2024improved,bai2025qwen3,shao2026gradient} have become increasingly capable general-purpose multimodal systems that support visual tasks within a unified generative interface~\cite{shao2025eventvad}. However, in practical deployments, a single pretrained model is rarely sufficient for downstream tasks. A solution is to train LoRA adapters for tasks \cite{hu2022lora,ijcai2025p683}, yielding a bank of adapters specialized for different domains~\cite{huang2023lorahub,shao2026tr,jiang2026medical,Yan_2026_WACV,yan2025renderworld}. This adapter-bank paradigm is parameter-efficient, but it introduces an inference-time challenge: \textbf{Given a LoRA weight from a target task, how should the model adapt to the target task with less forgetting?}

Existing adapter usage strategies are typically static~\cite{ilharco2022editing,yu2024language,du2024parameter,yadav2023ties,yang2024adamerging,shao2026icm,zeng2025robustmerge,padkan2025evaluating}. Activating only the target-task LoRA is safe but ignores transferable knowledge from related adapters, while fixed or similarity-based merging assigns the same adapter composition to all test inputs. Such task compositions cannot be adapted to instance-specific requirements, and direct parameter averaging may further distort the low-rank branch structure of LoRA adapters, causing unstable or harmful transfer~\cite{shao2025accidentblip,li2026near}. We therefore study label-free test-time LoRA fusion with a frozen backbone and a frozen bank of task-specific adapters. The goal is to construct a sample-specific fused adapter for each test input without updating either the backbone or the adapters. It is challenging because global fusion over the adapter bank can introduce negative transfer, and generative multimodal tasks do not share a unified objective for conventional test-time adaptation~\cite{sun2020test,zhang2022memo,niu2023towards,yuan2023robust,chen2026unpaired,wang2026unifying}.

To address these challenges, we propose \textbf{LiST}, a Local-Simplex Prompt-Level Test-Time LoRA Fusion framework. LiST first builds joint task representations by combining LoRA parameter anchors with prompt-level behavior vectors. For each target task, these representations are used to retrieve neighboring adapters and construct a target-conditioned local simplex, which restricts test-time search to a plausible task neighborhood. At inference time, LiST optimizes only low-dimensional simplex weights around a task-conditioned prior. Fusion is performed at the LoRA branch-output level rather than by averaging adapter parameters, preserving the structure of each task adapter. To select reliable weights without labels, LiST optimizes a prompt-level energy with prior, geometric, and stochastic-consistency constraints, followed by a safe acceptance rule that falls back to the task-conditioned prior when the searched solution is unreliable.

Empirical results show that LiST outperforms LoRA merging and conventional test-time adaptation baselines on multimodal and language tasks, while improving robustness on unseen tasks. 

Our contributions are summarized as follows:

\begin{itemize}
    \item We present \textbf{LiST}, a LoRA fusion method as label-free, sample-wise test-time fusion over a target-conditioned local simplex, optimizing only low-dimensional fusion weights while keeping the backbone frozen.

    \item We design a branch-preserving fusion and prompt-level energy framework that combines task uncertainty, stochastic consistency, target prior, and geometric constraints, together with a safe fallback rule for reliable adaptation.

    \item We demonstrate strong results on multimodal and language benchmarks, where LiST outperforms merging and conventional TTA baselines while improving unseen-task robustness.
\end{itemize}

\section{Related Work}
\subsection{Model Fusion}
Low-rank Adaptation (LoRA)~\cite{hu2022lora} is now a standard fine-tuning paradigm. In multi-task settings, early work viewed fine-tuned modules as parameter-space directions and combined them linearly~\cite{ilharco2022editing,liao2025gm}, but naive merging can cause task interference, leading to conflict-aware or component-filtering methods~\cite{yu2024language,yadav2023ties}. Adaptive approaches such as AdaMerging~\cite{yang2024adamerging} learn task- or layer-wise coefficients from unlabeled data, while PEFT-oriented methods further improve LoRA fusion, including PCB-merging~\cite{du2024parameter}, LoraHub~\cite{huang2023lorahub}, RobustMerge~\cite{zeng2025robustmerge}, and ICM-Fusion~\cite{shao2026icm}. Recent methods make fusion input-dependent by predicting example-conditioned weights~\cite{qorbani2025semantic}, including extensions to multimodal PEFT~\cite{guo2025hide}. Unlike these mostly static fusion approaches, our method treats LoRA composition as a continuous test-time search over an adapter manifold, yielding training-free, sample-specific compositions.

\subsection{Test-Time Adaptation}
Test-time adaptation improves robustness under distribution shift by refining models during inference. Early methods update model parameters using entropy minimization, self-supervision, or consistency objectives, such as Tent~\cite{wang2020tent}, Test-Time Training~\cite{sun2020test}, and MEMO~\cite{zhang2022memo}; later work also considers constrained settings such as single-instance adaptation~\cite{khurana2021sita}. While effective, these approaches typically adapt model parameters or high-dimensional states. Recent methods reduce the adaptation space: ELaTTA~\cite{luo2025efficient} searches over low-dimensional latent coefficients, and distribution-regularized refinement~\cite{chen2025test} constrains optimization to training-supported regions. Building on this shift from parameter updates to controlled search, our method adapts the composition of frozen task-specific LoRA adapters instead of latent states or predictions. This yields parameter-free, lightweight test-time adaptation compatible with multi-task adapter reuse, while enabling sample-specific refinement.


\section{Methodology}
\label{sec:method}
\begin{figure*}
    \centering
    \includegraphics[width=0.9\linewidth]{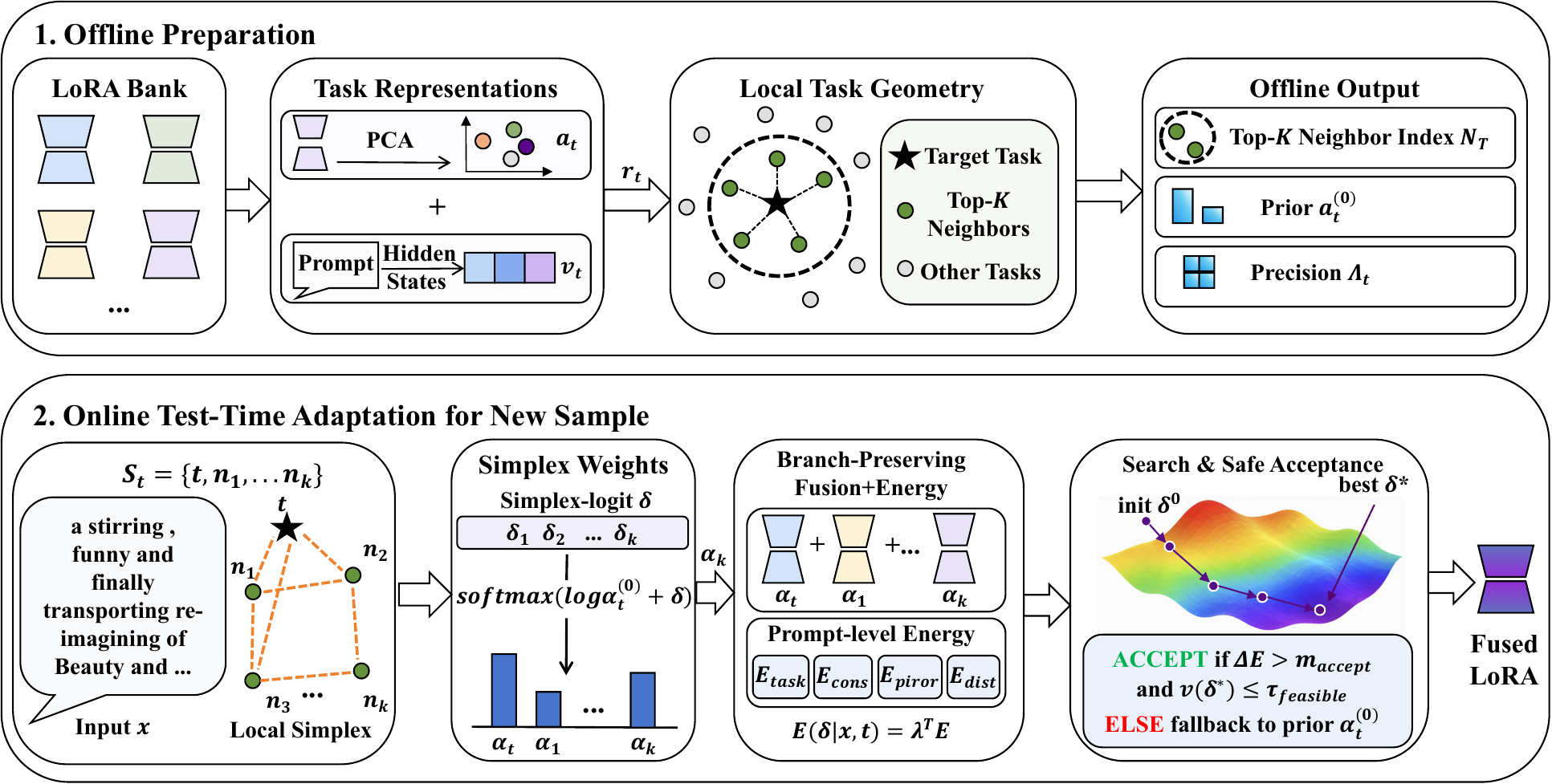}
    \caption{\textbf{Pipeline of LiST.} LiST first builds a local task simplex from an existing LoRA bank using task representations and neighbor retrieval. At test time, it searches simplex weights for each input, evaluates prompt-level energy under branch-preserving LoRA fusion, and accepts the searched weights only when they improve energy while satisfying feasibility constraints; otherwise, it falls back to the task-conditioned prior.}
    \label{fig:placeholder}
\end{figure*}
\subsection{Problem and Overview}

We study label-free test-time LoRA fusion with a frozen backbone and a frozen bank of task-specific adapters. Let $f_{\theta_0}$ denote the pretrained backbone and let
\begin{equation}
\mathcal{A}=\{\Delta\theta_1,\Delta\theta_2,\ldots,\Delta\theta_N\}
\label{eq:lora_bank}
\end{equation}
denote the LoRA bank, where $\Delta\theta_t$ is the adapter trained for task $t$. Given an unlabeled test input $x$ from target task $t$, LiST predicts a fused adapter parameterized by final simplex weights $\alpha^\dagger(x,t)$:
\begin{equation}
\hat{y}=f_{\theta_0,\alpha^\dagger(x,t)}(x).
\label{eq:prediction}
\end{equation}
All LoRA weights remain fixed. The only online variable is a low-dimensional simplex-logit perturbation $\delta$, which determines the fusion weights.

LiST separates offline task-level preparation from online sample-level adaptation. Offline, it builds task representations and stores four reusable objects for each target task: the local neighbor set $\mathcal{S}_t$, the task-conditioned prior $\alpha_t^{(0)}$, the local precision matrix $\Lambda_t$, and the original LoRA branches. Online, these objects are fixed, and LiST searches only $\delta$ for each test input. The searched perturbation $\delta^\star$ and the final deployed weight $\alpha^\dagger$ are defined in Section~\ref{sec:safe_prediction}.

\subsection{Local Simplex Construction}

LiST constructs a target-conditioned local simplex from two complementary task descriptors. First, for each task $t$, we flatten all trainable LoRA tensors into a task-level parameter vector $\ell_t$:
\begin{equation}
\ell_t=\operatorname{concat}\big(
\operatorname{vec}(W_{t,1}),\ldots,\operatorname{vec}(W_{t,M})
\big)\in\mathbb{R}^{D}.
\label{eq:lora_vector}
\end{equation}
Then it is projected into a low-dimensional anchor space:
\begin{equation}
a_t=U^\top \ell_t\in\mathbb{R}^{d_a},
\qquad d_a\ll D,
\label{eq:lora_anchor}
\end{equation}
where $U$ is obtained by PCA over the LoRA bank. The anchor $a_t$ is used only for neighbor retrieval and geometric regularization; the original LoRA parameters are never replaced.

Second, parameter proximity alone may not capture functional task similarity, so we extract a prompt-level behavior vector. Given a small support set $Q_t$, we run the backbone with adapter $\Delta\theta_t$ and average hidden representations:
\begin{equation}
v_t=\frac{1}{|Q_t|}\sum_{x\in Q_t} h_t(x).
\label{eq:behavior_vector}
\end{equation}

We optionally project the behavior vector as $\tilde v_t=P(v_t)$ and then concatenate it with the anchor to obtain the final task representation $r_t=[a_t;\tilde v_t]$.

We compute task similarity by cosine similarity:
\begin{equation}
\operatorname{sim}(t,i)=
\frac{r_t^\top r_i}{\|r_t\|_2\|r_i\|_2}.
\label{eq:task_similarity}
\end{equation}
For target task $t$, we retrieve top-$K$ neighboring adapters and define the local simplex as
\begin{equation}
\mathcal{S}_t=
\{t\}\cup
\operatorname{TopK}_{i\neq t}\operatorname{sim}(t,i).
\label{eq:local_simplex}
\end{equation}
Thus, online search is restricted to the target adapter and a small set of related adapters.

Inside $\mathcal{S}_t$, LiST defines a task-conditioned prior $\alpha_t^{(0)}$. The target LoRA receives mass $\rho$ as $\alpha_{t,t}^{(0)}=\rho$.

The remaining mass is distributed to neighbors according to similarity:
\begin{equation}
\alpha_{t,j}^{(0)}
=
(1-\rho)
\frac{\exp(\operatorname{sim}(t,j)/\tau)}
{\sum_{k\in\mathcal{S}_t\setminus\{t\}}
\exp(\operatorname{sim}(t,k)/\tau)},
\quad j\neq t.
\label{eq:neighbor_prior}
\end{equation}
We also estimate the local covariance of anchors in $\mathcal{S}_t$:
\begin{equation}
\Sigma_t=
\operatorname{Cov}\{a_i\mid i\in\mathcal{S}_t\}
+\epsilon I,
\label{eq:local_covariance}
\end{equation}
and denote precisely with $\Lambda_t=\Sigma_t^\dagger$.

The prior $\alpha_t^{(0)}$ is used both as the initialization of online search and as the fallback solution. The precision matrix $\Lambda_t$ defines a local trust region by measuring how far the fused anchor barycenter deviates from the target anchor.

\paragraph{Unseen-domain inference.}
For unseen-domain evaluation, no target-domain LoRA adapter is trained or
used. LiST therefore forms the local simplex from the retrieved seen-domain
adapters only, and replaces the task-conditioned prior in Eq.~(10)--(11)
with a similarity-normalized prior:
\begin{equation}
\alpha^{(0)}_{j}
=
\frac{\exp(\mathrm{sim}(u,j)/\tau)}
{\sum_{k \in \mathcal{S}} \exp(\mathrm{sim}(u,k)/\tau)},
\quad j \in \mathcal{S}.
\end{equation}
The same search objective and acceptance rule are applied, with fallback
to this retrieval-based prior.
\subsection{Energy Aware Branch-Preserving Fusion}

LiST avoids direct LoRA parameter averaging. For a LoRA-augmented linear layer with frozen base weight $W_0$, each adapter $i\in\mathcal{S}_t$ has low-rank branch $(A_i,B_i)$ and scale $s_i$. Given simplex weights $\alpha$, LiST fuses branch outputs as
\begin{equation}
y=
W_0x+
\sum_{i\in\mathcal{S}_t}
\alpha_i s_i B_i(A_i x).
\label{eq:branch_fusion}
\end{equation}
This preserves the structure of every LoRA branch and reduces online adaptation to simplex-weight optimization.

We parameterize simplex weights by an unconstrained perturbation $\delta$ around the task-conditioned prior:
\begin{equation}
\alpha(\delta)=
\operatorname{softmax}
\big(
\log\alpha_t^{(0)}+\delta
\big).
\label{eq:simplex_logit}
\end{equation}
When $\delta=0$, the fusion weight recovers the prior:
\begin{equation}
\alpha(0)=\alpha_t^{(0)}.
\label{eq:prior_recovery}
\end{equation}

For each candidate $\delta$, LiST performs $M$ stochastic forward passes and obtains predictive objects in the model's native prediction space:
\begin{equation}
\{q_m(x,\delta)\}_{m=1}^{M}.
\label{eq:predictive_objects}
\end{equation}
The main task energy is
\begin{equation}
E_{\mathrm{task}}(\delta|x)
=
\frac{1}{M}
\sum_{m=1}^{M}
U(q_m(x,\delta)),
\label{eq:task_energy}
\end{equation}
where $U(\cdot)$ is a task-aware uncertainty functional.

To avoid unreliable low-uncertainty solutions, LiST adds three feasibility terms. The prior deviation is
\begin{equation}
E_{\mathrm{prior}}(\delta,t)
=
\operatorname{KL}
\big(
\alpha(\delta)\|
\alpha_t^{(0)}
\big).
\label{eq:prior_energy}
\end{equation}
The fused anchor barycenter is
\begin{equation}
z(\delta)=
\sum_{i\in\mathcal{S}_t}
\alpha_i(\delta)a_i.
\label{eq:barycenter}
\end{equation}
The local geometric distance is
\begin{equation}
E_{\mathrm{dist}}(\delta,t)
=
(z(\delta)-a_t)^\top
\Lambda_t
(z(\delta)-a_t).
\label{eq:distance_energy}
\end{equation}
When predictive objects admit distributions, stochastic consistency is measured by
\begin{equation}
E_{\mathrm{cons}}(\delta|x)
=
\operatorname{JS}_{\mathrm{pairwise}}
(q_1,\ldots,q_M).
\label{eq:consistency_energy}
\end{equation}
where $JS$ indicates Jensen-Shannon Entropy.

Since these terms have different scales, LiST uses a relative normalized augmented-Lagrangian objective. We normalize the task energy against the prior baseline:
\begin{equation}
\widetilde E_{\mathrm{task}}
=
\frac{
E_{\mathrm{task}}(\delta|x,t)
-
E_{\mathrm{task}}(0|x,t)
}{
s_{\mathrm{task}}+\epsilon
}.
\label{eq:normalized_task_energy}
\end{equation}
Each constraint is converted into a normalized violation:
\begin{equation}
c_i=
\frac{E_i-\tau_i}{s_i+\epsilon},
\qquad
i\in\{\mathrm{prior},\mathrm{dist},\mathrm{cons}\}.
\label{eq:constraint_violation}
\end{equation}
The final energy is
\begin{equation}
\mathcal{E}(\delta|x,t)
=
\widetilde E_{\mathrm{task}}
+
\sum_i
\lambda_i[c_i]_+
+
\frac{\rho_{\mathrm{AL}}}{2}
\sum_i[c_i]_+^2,
\label{eq:final_energy}
\end{equation}
where $[u]_+=\max(u,0)$.

If the consistency term is undefined for a task-specific predictive object, it is omitted.

Because each energy component is normalized relative to the prior baseline or converted into a normalized violation, LiST uses fixed coefficients across tasks. The thresholds and normalization scales are estimated from offline task statistics, and the global hyperparameters are kept fixed throughout evaluation. Sensitivity analyses for $K$, $M$, and $m_{\mathrm{accept}}$ are reported in Section~\ref{sec:ablation}.

\begin{table*}[t]
\centering
\scriptsize
\setlength{\tabcolsep}{3.2pt}
\renewcommand{\arraystretch}{0.92}
\resizebox{\textwidth}{!}{%
\begin{tabular}{l|ccccccccc|ccccc}
\toprule
\multicolumn{1}{c|}{} 
& \multicolumn{9}{c|}{\textsc{Seen Tasks}} 
& \multicolumn{5}{c}{\textsc{Unseen Tasks}} \\
\cmidrule(lr){2-10}
\cmidrule(lr){11-15}
Method 
& SciQA & Image & VQA & REC & OCR & VizWiz & Flickr & IconQA & \textbf{Avg.}
& AVQA & Image-R & S2W & TabMWP & \textbf{Avg.} \\
\midrule

\rowcolor{blue!10}
\multicolumn{15}{c}{\textbf{Results on Qwen3-VL-8B}} \\
\midrule

Original Model
& 85.97 & 35.01 & 63.37 & 48.13 & 62.91 & 27.26 & 28.63 & 67.50 & 52.35
& 81.77 & 81.47 & 10.82 & 58.30 & 58.09 \\

LoRA
& 97.71 & 97.74 & \textbf{75.22} & 89.21 & 72.83 & \textbf{74.94} & 61.69 & \textbf{99.16} & 83.56 
& -- & -- & -- & -- & -- \\

\midrule

Task-Arithmetic~\cite{ilharco2022editing}
& 78.97 & 73.44 & 60.53 & 78.05 & 62.85 & 45.34 & 38.17 & 88.79 & 65.77
& 69.30 & 80.72 & 10.86 & 61.23 & 55.52  \\

TIES-Merging~\cite{yadav2023ties}
& 75.05 & 81.17 & 64.75 & 63.92 & 61.66 & 41.84 & 39.06 & 84.25 & 63.96
& 67.86 & 82.89 & 9.85 & 65.11 & 56.43 \\

DARE~\cite{yu2024language}
& 87.36 & 81.70 & 60.33 & 73.32 & 54.59 & 48.86 & 38.79 & 88.06 & 66.63
& 68.29 & 83.01 & 9.38 & 65.35 & 56.50 \\

PCB-Merging~\cite{du2024parameter}
& 83.61 & 47.25 & 53.65 & 56.21 & 59.50 & 33.76 & 28.03 & 73.09 & 54.39 
& 71.73 & 85.12 & 9.64 & 64.87 & 57.84 \\

ICM-Fusion~\cite{shao2026icm}
& 91.75 & 86.98 & 67.48 & 77.81 & 64.10 & 53.23 & 43.48 & 89.69 & 71.81 
& 75.72 & 84.38 & 8.33 & 68.76 & 59.29 \\

RobustMerge~\cite{zeng2025robustmerge}
& 87.31 & 81.56 & 64.16 & 77.20 & 71.45 & 48.99 & 41.72 & 91.97 & 70.55
& 68.29 & 84.91 & 11.10 & 65.89 & 57.54  \\

AdaMerging~\cite{yang2024adamerging}
& 83.35 & 40.28 & 63.31 & 74.72 & 58.46 & 41.74 & 34.82 & 71.40 & 58.51
& 73.91 & 85.47 & 8.64 & 66.29 & 58.58 \\

TENT~\cite{wang2020tent}
& 96.67 & 95.82 & 72.07 & 87.97 & 69.28 & 72.13 & 56.92 & 98.09 & 81.12 
& 81.67 & 87.17 & 12.46 & 71.15 & 63.11  \\

MEMO~\cite{zhang2022memo}
& 96.64 & 96.82 & 71.94 & 86.03 & 71.91 & 71.91 & 59.84 & 96.86 & 81.49 
& 83.85 & 88.45 & 13.19 & 71.74 & 64.31 \\

\midrule

\textbf{LiST (Ours)}
& \textbf{97.75} & \textbf{97.78} & 75.16 & \textbf{90.03} & \textbf{73.12} & 74.88 & \textbf{62.88} & 99.09 & \textbf{83.84}
& \textbf{87.68} & \textbf{89.94} & \textbf{16.66} & \textbf{78.42} & \textbf{68.18} \\

\midrule
\rowcolor{blue!10}
\multicolumn{15}{c}{\textbf{Results on LLaVA-V1.5-7B}} \\
\midrule

Original Model
& 71.73 & 40.87 & 42.88 & 36.10 & 31.23 & 41.03 & 39.07 & 38.09 & 42.62
& 78.45 & 70.36 & 10.05 & 23.10 & 45.49 \\

LoRA
& 90.73 & 97.40 & \textbf{68.12} & 75.70 & \textbf{54.55} & 70.12 & 57.28 & \textbf{73.33} & 73.40
& -- & -- & -- & -- & -- \\

\midrule

Task-Arithmetic~\cite{ilharco2022editing}
& 70.98 & 57.43 & 53.89 & 35.66 & 35.31 & 46.82 & 42.56 & 36.68 & 47.42 
& 67.03 & 68.82 & 8.46 & 33.34 & 44.41 \\

TIES-Merging~\cite{yadav2023ties}
& 68.94 & 58.92 & 49.18 & 36.19 & 31.28 & 41.98 & 42.41 & 31.53 & 45.05 
& 66.94 & 67.57 & 9.72 & 34.14 & 44.59 \\

DARE~\cite{yu2024language}
& 68.76 & 57.63 & 52.84 & 32.84 & 36.18 & 39.18 & 45.77 & 37.67 & 46.36 
& 68.25 & 66.94 & 8.39 & 33.78 & 44.34 \\

PCB-Merging~\cite{du2024parameter}
& 70.63 & 61.89 & 51.22 & 36.73 & 35.99 & 42.39 & 43.12 & 31.98 & 46.74 
& 69.30 & 66.17 & 9.46 & 33.21 & 44.53 \\

ICM-Fusion~\cite{shao2026icm}
& 78.94 & 75.32 & 62.41 & 45.68 & 41.96 & 57.15 & 44.17 & 59.34 & 58.12  
& 76.52 & 69.54 & 9.84 & 35.68 & 47.90 \\

RobustMerge~\cite{zeng2025robustmerge}
& 73.43 & 65.54 & 56.81 & 42.84 & 38.77 & 46.61 & 33.81 & 45.82 & 50.45
& 68.77 & 67.98 & 8.84 & 32.82 & 44.60 \\

AdaMerging~\cite{yang2024adamerging}
& 71.23 & 71.37 & 47.21 & 38.19 & 30.41 & 47.85 & 41.92 & 42.43 & 48.83
& 69.38 & 66.43 & 7.29 & 32.20 & 43.83 \\

TENT~\cite{wang2020tent}
& 82.65 & 95.87 & 63.07 & 63.68 & 43.67 & 58.84 & 46.97 & 60.23 & 64.37 
& 78.41 & 78.92 & 12.26 & 42.18 & 52.94 \\

MEMO~\cite{zhang2022memo}
& 84.01 & 96.35 & 65.91 & 65.93 & 42.58 & 61.88 & 47.19 & 59.18 & 65.38
& 79.57 & 77.23 & 13.48 & 43.23 & 53.38 \\

\midrule

\textbf{LiST (Ours)}
& \textbf{91.65} & \textbf{97.59} & 67.99 & \textbf{75.73} & 54.52 & \textbf{70.16} & \textbf{57.37} & 73.26 & \textbf{73.53}
& \textbf{83.68} & \textbf{81.94} & \textbf{15.34} & \textbf{46.64} & \textbf{56.15} \\

\bottomrule
\end{tabular}%
}
\caption{\textbf{Multimodal results.} 
Across seen and unseen tasks, LiST delivers the best average performance across multimodal backbones, preserving task-specific LoRA utility while improving generalization to unseen distributions.}
\label{tab:multimodal_results_combined}
\end{table*}
\subsection{Search and Safe Prediction}
\label{sec:safe_prediction}

The objective $\mathcal{E}(\delta|x,t)$ is stochastic and derivative-free, but the search dimension is only $|\mathcal{S}_t|$. We therefore use a standard two-stage CMA-ES optimizer~\cite{luo2025efficient} over the simplex-logit variable $\delta$, initialized at $\delta_0=0$

The best searched perturbation is denoted by
\begin{equation}
\delta^\star=
\arg\min_{\delta}
\mathcal{E}(\delta|x,t).
\label{eq:best_delta}
\end{equation}

LiST does not always deploy the lowest-energy candidate. Let the energy improvement over the prior be
\begin{equation}
\Delta E=
\mathcal{E}(0|x,t)
-
\mathcal{E}(\delta^\star|x,t),
\label{eq:energy_improvement}
\end{equation}
and let the maximum normalized violation be
\begin{equation}
\nu(\delta^\star)=
\max
\{
c_{\mathrm{prior}}(\delta^\star),
c_{\mathrm{dist}}(\delta^\star),
c_{\mathrm{cons}}(\delta^\star)
\}.
\label{eq:max_violation}
\end{equation}
LiST accepts the searched candidate only if
\begin{equation}
\Delta E>m_{\mathrm{accept}}
\quad\text{and}\quad
\nu(\delta^\star)\le \tau_{\mathrm{feasible}}.
\label{eq:acceptance_rule}
\end{equation}
The final deployed fusion weight is
\begin{equation}
\alpha^\dagger(x,t)=
\begin{cases}
\alpha(\delta^\star), & \text{if Eq.~\ref{eq:acceptance_rule} holds},\\
\alpha_t^{(0)}, & \text{otherwise}.
\end{cases}
\label{eq:final_weight}
\end{equation}
The final prediction is generated by the frozen backbone with $\alpha^\dagger$. This safe rule prevents high-violation low-energy candidates from being deployed and connects online search to the offline task-conditioned prior.

\section{Experiment}
\subsection{Experimental Setup}
\paragraph{Training Details.}
For multimodal tasks, we perform experiments on Qwen3-VL-8B~\cite{bai2025qwen3} and LLaVA-V1.5-7B~\cite{liu2024improved}. For language tasks, we perform experiments on DeepSeek-7B~\cite{bi2024deepseek} and Qwen3-8B~\cite{yang2025qwen3}. We perform all experiments on four RTX 4090 GPUs.

\paragraph{Benchmarks.}
We widely evaluate multimodal tasks on MM-MergeBench~\cite{zeng2025robustmerge} and language tasks on GLUE benchmark~\cite{wang2018glue}. Details can be found in Sec.~\ref{benchmark}.

\paragraph{Compared Baselines.} 
We extensively compare our method against model fusion and test-time adaptation (TTA) methods to demonstrate its effectiveness. Specifically, the model fusion methods include Task-Arithmetic~\cite{ilharco2022editing}, TIES-Merging~\cite{yadav2023ties}, DARE~\cite{yu2024language}, PCB-Merging~\cite{du2024parameter}, ICM-Fusion~\cite{shao2026icm} and RobustMerge~\cite{zeng2025robustmerge}. The TTA method includes AdaMerging~\cite{yang2024adamerging}, MEMO~\cite{zhang2022memo} and TENT~\cite{wang2020tent}.

\subsection{Main Results}

\begin{table*}[t]
\centering
\resizebox{\textwidth}{!}{%
\begin{tabular}{l|ccccccccccc}
\toprule
Method & SST-2 & MRPC & RTE & QNLI & QQP & CoLA & MNLI-m & MNLI-mm & STS-B & WNLI & Average \\
\midrule

\rowcolor{blue!10}
\multicolumn{12}{c}{\textbf{Results on DeepSeek-7B}} \\
\midrule

Original Model
& 60.89 & 58.03 & 47.65 & 52.11 & 40.13 & 7.85 & 33.70 & 34.02 & 13.87 & 47.88 & 39.61 \\

LoRA
& 97.24 & \best{88.72} & \best{89.16} & 94.96 & 90.47 & 67.49 & 91.12 & 91.02 & 90.16 & 77.46 & 87.78 \\

\midrule

Task-Arithmetic~\cite{ilharco2022editing}
& 58.37 & 35.47 & 58.48 & 40.14 & 36.83 & 8.09 & 58.75 & 60.15 & 24.13 & 57.74 & 43.81 \\

TIES-Merging~\cite{yadav2023ties}
& 68.92 & 48.69 & 68.92 & 51.05 & 47.77 & 43.65 & 46.86 & 49.73 & 34.01 & 54.36 & 51.40 \\

DARE~\cite{yu2024language}
& 49.08 & 15.80 & 47.29 & 52.26 & 39.41 & 4.69 & 32.89 & 33.45 & 21.03 & 56.33 & 35.22 \\

PCB-Merging~\cite{du2024parameter}
& 75.57 & 49.56 & 46.21 & 51.19 & 47.71 & 37.09 & 41.79 & 44.61 & 24.46 & 47.88 & 46.61 \\

ICM-Fusion~\cite{shao2026icm}
& 82.19 & 78.76 & 72.53 & 69.21 & 63.68 & 52.84 & 76.69 & 78.13 & 63.18 & 61.17 & 69.84 \\

RobustMerge~\cite{zeng2025robustmerge}
& 72.82 & 41.96 & 63.17 & 51.27 & 33.86 & 28.14 & 79.93 & 81.20 & 46.85 & 56.33 & 55.55 \\

AdaMerging~\cite{yang2024adamerging}
& 60.43 & 50.82 & 61.73 & 50.85 & 35.35 & 24.74 & 72.26 & 72.79 & 59.72 & 45.07 & 53.38 \\

TENT~\cite{wang2020tent}
& 93.91 & 85.29 & 86.08 & 89.87 & 89.16 & 62.91 & 87.17 & 88.21 & 86.09 & 74.78 & 84.35 \\

MEMO~\cite{zhang2022memo}
& 94.76 & 85.74 & 85.32 & 91.89 & 87.97 & 65.38 & 89.08 & 88.93 & 89.73 & 75.12 & 85.39 \\

\midrule

\best{LiST (Ours)}
& \best{97.25} & 88.63 & 88.95 & \best{94.98} & \best{90.49} & \best{68.24} & \best{91.16} & \best{91.07} & \best{90.18} & \best{77.84} & \best{87.88} \\

\midrule
\rowcolor{blue!10}
\multicolumn{12}{c}{\textbf{Results on Qwen3-8B}} \\
\midrule

Original Model
& 74.03 & 64.61 & 66.64 & 70.48 & 51.83 & 42.89 & 51.02 & 49.96 & 64.45 & 39.43 & 57.53 \\

LoRA
& 96.67 & 89.97 & 89.89 & 95.78 & 90.70 & 65.94 & 91.63 & 91.62 & 90.08 & 70.42 & 87.27 \\

\midrule

Task-Arithmetic~\cite{ilharco2022editing}
& 86.90 & 68.35 & 79.13 & 75.69 & 60.80 & 54.85 & 74.12 & 74.44 & 78.77 & 56.33 & 70.94 \\

TIES-Merging~\cite{yadav2023ties}
& 65.36 & 74.91 & 48.01 & 73.67 & 80.01 & 52.64 & 57.20 & 64.91 & 51.96 & 57.74 & 62.64 \\

DARE~\cite{yu2024language}
& 76.33 & 69.25 & 69.89 & 72.71 & 57.21 & 42.37 & 81.47 & 81.26 & 70.10 & 47.60 & 66.82 \\

PCB-Merging~\cite{du2024parameter}
& 50.57 & 74.21 & 48.73 & 51.25 & 53.55 & 36.08 & 41.85 & 43.31 & 76.91 & 57.74 & 53.42 \\

ICM-Fusion~\cite{shao2026icm}
& 82.23 & 78.46 & 73.53 & 86.80 & 81.13 & 53.67 & 76.70 & 79.44 & 82.56 & 60.56 & 75.51 \\

RobustMerge~\cite{zeng2025robustmerge}
& 80.64 & 78.50 & 79.16 & 84.32 & 82.39 & 59.70 & 81.12 & 81.27 & 80.15 & 57.74 & 76.50 \\

AdaMerging~\cite{yang2024adamerging}
& 53.89 & 75.04 & 48.73 & 51.61 & 54.03 & 32.13 & 46.82 & 48.83 & 68.43 & 53.52 & 53.30 \\

TENT~\cite{wang2020tent}
& 85.73 & 81.76 & 80.48 & 91.59 & 84.82 & 60.23 & 81.28 & 81.47 & 84.45 & 63.38 & 79.52 \\

MEMO~\cite{zhang2022memo}
& 91.58 & 82.22 & 83.17 & 90.26 & 85.58 & 61.10 & 83.55 & 84.16 & 86.49 & 60.56 & 80.87 \\

\midrule

\best{LiST (Ours)}
& \best{96.79} & \best{90.27} & \best{90.25} & \best{95.84} & \best{90.82} & \best{67.35} & \best{91.73} & \best{91.80} & \best{90.31} & \best{71.83} & \best{87.70} \\

\bottomrule
\end{tabular}%
}
\caption{\textbf{Language results.} Across diverse tasks and LLM backbones, LiST achieves the best average performance, preserving task-specific understanding while leveraging transferable knowledge across related tasks.}
\label{tab:glue_results_combined}
\end{table*}

\begin{table}[t]
\centering
\begin{tabular}{l|c}
\toprule
\textbf{Method} & \textbf{Avg} \\
\midrule
Target LoRA & 87.35 \\
Static Prior Fusion & 78.83 \\
+ Energy Search & 82.75 \\
+ Prior Constraint & 83.32 \\
+ Prior + Geo & 85.91 \\
+ Prior + Geo + Stability & 87.55 \\
\midrule
\textbf{Full LiST} & \textbf{87.70} \\
\bottomrule
\end{tabular}
\vspace{-0.6em}
\caption{\textbf{Component ablation results.} }
\label{tab:component_ablation}
\end{table}

\begin{table}[t]
\centering
\resizebox{0.5\textwidth}{!}{%
\begin{tabular}{l|cccccc}
\toprule
\textbf{Variant} & \textbf{Avg} & \textbf{Entropy $\downarrow$} & \textbf{JS $\downarrow$} & \textbf{KL $\downarrow$} 
& \textbf{Geo. Dist. $\downarrow$} & \makecell{\textbf{Wrong}\\\textbf{Confident $\downarrow$}}  \\
\midrule
Entropy Only 
& 79.36 & 7.89 & 0.239 & 3.98 & 3.91 & 9.616 \\

Entropy + JS 
& 81.14 & 7.85 & 0.216 & 3.70 & 3.58 & 9.509 \\

Entropy + Prior 
& 83.97 & 7.73 & 0.174 & 3.118 & 2.953 & 9.505 \\

Entropy + Geo 
& 86.12 & 7.43 & 0.163 & 2.311 & 1.011 & 9.478 \\

Prior + Geo 
& 86.69 & 7.35 & 0.101 & 1.684 & 0.859 & 9.450 \\

\midrule
\textbf{Full Energy} 
& \textbf{87.70} & \textbf{7.13} & \textbf{0.078} & \textbf{1.178} & \textbf{0.772} & \textbf{9.227} \\
\bottomrule
\end{tabular}%
}
\vspace{-0.6em}
\caption{\textbf{Energy and constraint ablation results.}}
\label{tab:energy_constraint_ablation}
\end{table}

\begin{table}[t]
\centering
\resizebox{0.45\textwidth}{!}{%
\begin{tabular}{l|cccc}
\toprule
\makecell{\textbf{Neighbor}\\\textbf{Strategy}} & Avg 
& \makecell{\textbf{Static}\\\textbf{Fusion}} 
& \makecell{\textbf{Neg.}\\\textbf{Transfer} $\downarrow$} 
& \makecell{\textbf{Geo.}\\\textbf{Dist.} $\downarrow$} \\
\midrule
Target Only & 87.35 & 84.18 & \textbf{0.00} & \textbf{0.00} \\
Random K & 82.95 & 80.99 & 0.13 & 2.87 \\
Anchor Only & 84.97 & 85.01 & 0.11 & 0.90 \\
Behavior Only & 84.80 & 85.86 & 0.09 & 0.83 \\
Joint Representation & 86.22 & 86.12 & 0.07 & 0.74 \\
Global Simplex & 86.87 & 86.43 & 0.11 & 1.02 \\
\midrule
\textbf{Full Local Simplex} & \textbf{87.70} & \textbf{87.68} & 0.02 & 0.65 \\
\bottomrule
\end{tabular}}
\vspace{-0.6em}
\caption{\textbf{Ablation on local simplex construction.}}
\label{tab:local_simplex_ablation}
\end{table}

\begin{table*}[t]
\centering
\resizebox{0.8\textwidth}{!}{%
\begin{tabular}{l|ccccc}
\toprule
\textbf{Variant} & \textbf{Avg.} & \textbf{Accepted Subset} & \textbf{Rejected Subset} & \textbf{Neg. Transfer $\downarrow$} & \textbf{Fallback Rate} \\
\midrule
Prior Only 
& 80.84 & 76.09 & \textbf{88.41} & 0.17 & -- \\

Always Accept 
& 82.83 & 78.16 & 82.38 & 0.13 & 0.00 \\

Energy-only Accept 
& 86.60 & 79.38 & 84.57 & 0.09 & 29.71 \\

Constraint-only Accept 
& 86.92 & 80.79 & 86.06 & 0.08 & 36.38 \\

\midrule
\textbf{Full Accept Rule} 
& \textbf{87.70} & \textbf{82.76} & \textbf{88.41} & \textbf{0.02} & \textbf{59.92} \\
\bottomrule
\end{tabular}%
}
\vspace{-0.6em}
\caption{\textbf{Safe fallback ablation results.}}
\label{tab:safe_fallback_ablation}
\end{table*}

\paragraph{Outperforming Across Multimodal Tasks.}
As shown in Table~\ref{tab:multimodal_results_combined}, on both models, LiST achieves the best overall performance across multimodal benchmarks. It consistently improves over the original model and remains close to, or even surpasses, single-task LoRA on several datasets, indicating effective use of transferable knowledge while preserving target-task capability.
Compared with static merging methods and test-time adaptation baselines, LiST better balances fusion and adaptation. These results show that sample-adaptive local-simplex fusion can more effectively exploit complementary information among related LoRA adapters while preserving task-specific knowledge.

\paragraph{Consistent Effectiveness for Language Modality.}
As shown in Table~\ref{tab:glue_results_combined}, LiST also achieves strong performance on both models. It performs better than single-task LoRA on most tasks, showing that LiST can generalize to text-only LLM adaptation while selectively reusing useful cross-task knowledge.
It indicates LiST provides more stable and effective fusion. These results indicate that sample-wise test-time search better preserves target-task capability while exploiting the task-conditioned geometry of related LoRA adapters.

\paragraph{Out-of-Distribution Generalization. }
As shown in Table~\ref{tab:multimodal_results_combined}, we further evaluates LiST on unseen domains in MM-MergeBench.
These domains contain only test samples and provide no training samples
for constructing target-domain LoRA adapters, so no unseen-domain LoRA
is available during evaluation. Under this stricter protocol, LiST reuses
only the adapters trained on seen domains and constructs the local simplex
from retrieved seen-domain adapters. Despite the absence of target-domain
adapters, LiST achieves the best unseen-domain average on both Qwen3-VL-8B
and LLaVA-V1.5-7B, consistently improving over the original model as well
as static merging and test-time adaptation baselines. These results suggest
that sample-wise search over a local adapter simplex can transfer useful
capabilities from related seen domains, while the task-conditioned prior
and feasibility-aware acceptance rule help reduce negative transfer from
irrelevant adapters.

\subsection{Ablation Studies}
All ablation experiments are conducted on the GLUE benchmark using Qwen3-8B.
\subsubsection{Effect of Core Components}
As shown in Table~\ref{tab:component_ablation}, we ablate the major components of LiST. The target-task LoRA provides a strong and safe baseline but cannot exploit knowledge from related adapters. Static prior fusion performs worse, suggesting that fixed-weight adapter mixing can disrupt task-specific behavior. Test-time energy search improves performance through sample-wise adapter selection, while the prior constraint stabilizes the search by limiting deviation from the target task. The geometric constraint yields a larger gain, highlighting the importance of staying close to the local task manifold. Stochastic stability further improves robustness, and Full LiST achieves the best average result. Overall, these results show that LiST benefits from combining sample-wise search, target-aware prior regularization, local geometry, and stochastic consistency.

\subsubsection{Ablation on Energy Constraints}
Table~\ref{tab:energy_constraint_ablation} ablates the energy components. Entropy-only optimization gives the weakest performance and the highest wrong-confident score, confirming that uncertainty minimization alone can produce over-confident unreliable predictions. Adding stochastic consistency improves performance and JS stability by encouraging agreement under LoRA branch perturbations. Prior and geometric constraints further restrict the search to task-consistent and locally plausible regions, with the geometric term giving a particularly strong gain. The full energy achieves the best average performance and the lowest entropy, JS divergence, KL deviation, geometric distance, and wrong-confident score. These results show that the proposed energy is a feasibility-aware surrogate that balances confidence, stability, task identity, and local manifold plausibility.

\begin{figure*}[t]
    \centering
    \begin{subfigure}[t]{0.48\textwidth}
        \centering
        \includegraphics[width=\linewidth]{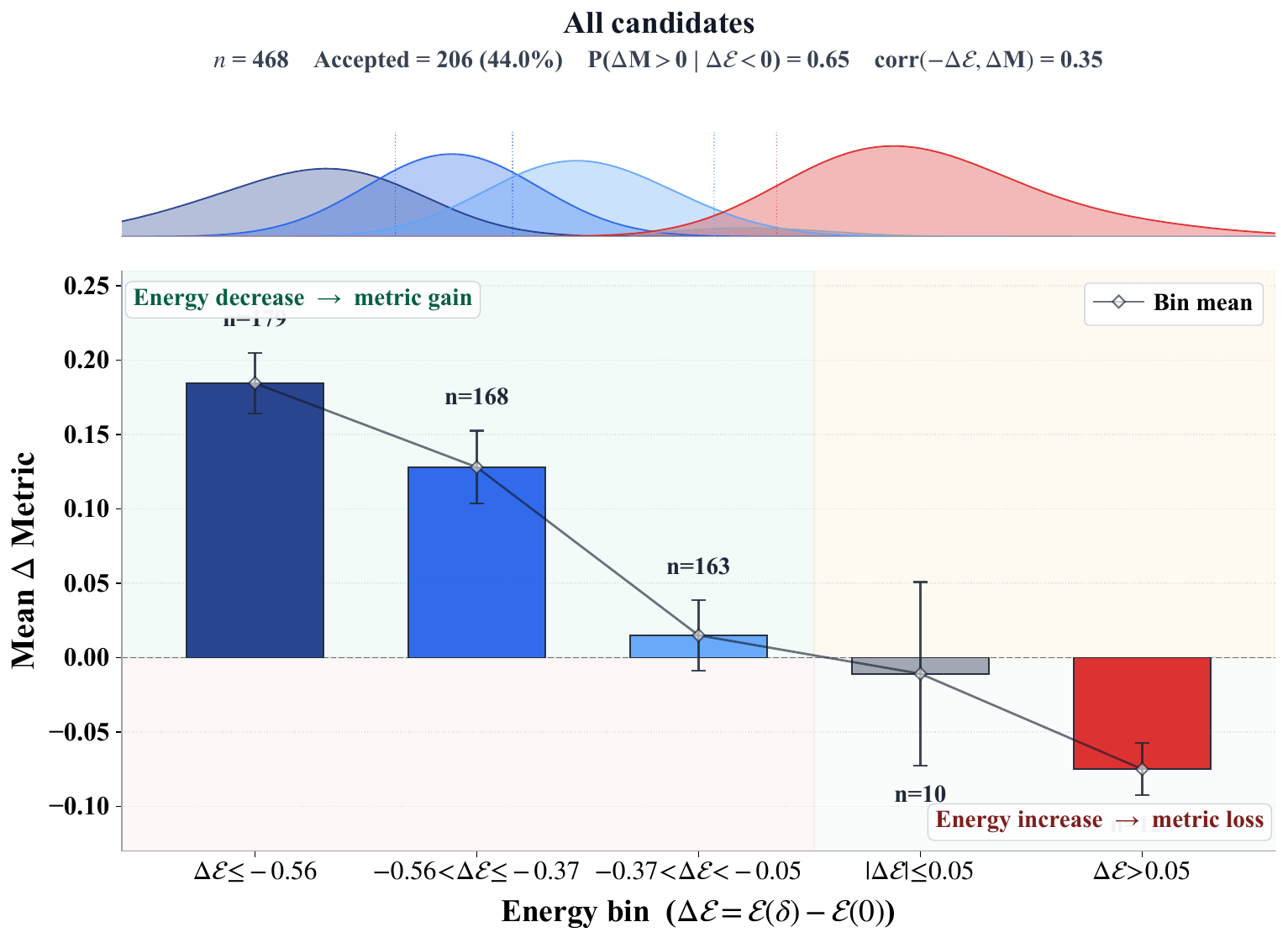}
        \caption{Binned energy change vs. metric gain.}
        \label{fig:energy_bin}
    \end{subfigure}
    \hfill
    \begin{subfigure}[t]{0.48\textwidth}
        \centering
        \includegraphics[width=\linewidth]{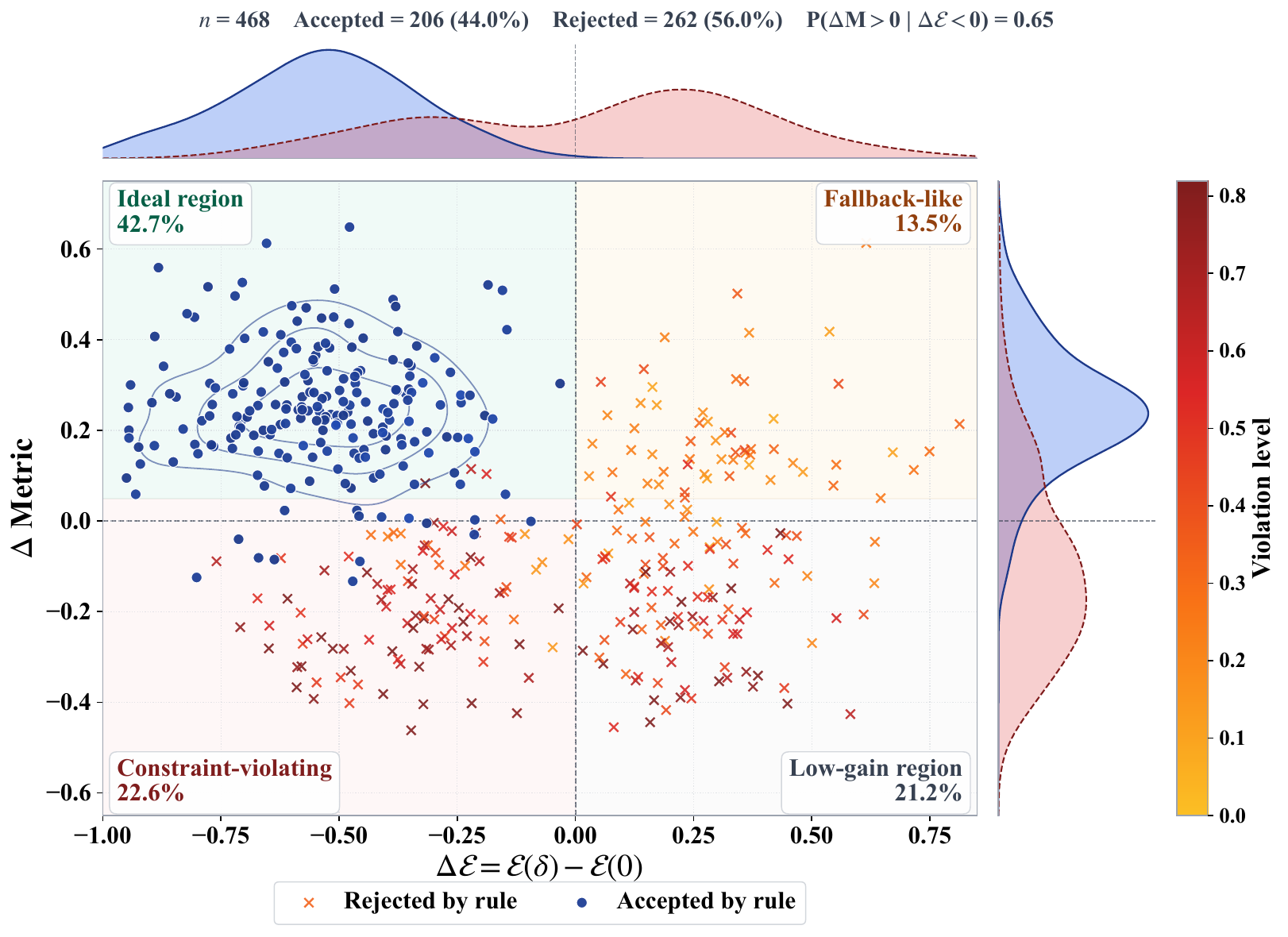}
        \caption{Candidate-level alignment under constraint violations.}
        \label{fig:scatter_violation}
    \end{subfigure}
    \caption{\textbf{Energy--performance alignment under feasibility constraints.}
Candidate perturbations are grouped by relative energy change, $\Delta\mathcal{E}=\mathcal{E}(\delta)-\mathcal{E}(0)$. Larger energy decreases generally correspond to larger downstream gains. However, low-energy candidates with high normalized constraint violations can still hurt performance, indicating that energy minimization must be coupled with feasibility-aware selection.
    }
    \label{fig:energy_alignment}
\end{figure*}

\subsubsection{Ablation on Local Simplex Construction}
Table~\ref{tab:local_simplex_ablation} compares different simplex construction strategies. The target-only baseline is safe but lacks cross-task transfer, while random neighbors cause strong negative transfer. Anchor-only and behavior-only retrieval improve over random selection but remain suboptimal, suggesting that parameter geometry and model behavior provide complementary task-similarity signals. Their joint representation further improves performance and reduces geometric displacement. Global simplex search underperforms the full local simplex, indicating that irrelevant adapters weaken the local manifold assumption. Full local simplex achieves the best average performance, low negative transfer, and small geometric distance, confirming the importance of target-conditioned local retrieval for reliable adapter reuse.

\subsubsection{Safe acceptance and fallback.}
Table~\ref{tab:safe_fallback_ablation} compares acceptance rules for deploying searched fusion weights. The prior-only baseline is stable but cannot fully exploit transferable adapters, while always accepting searched candidates improves performance but increases negative transfer. Energy-only and constraint-only rules help, but each misses part of the reliability criterion. The full acceptance rule achieves the best average performance and the lowest negative transfer by requiring both sufficient energy improvement and low constraint violation. Its higher fallback rate shows that LiST rejects risky candidates and returns to the task-conditioned prior when search is unreliable.
\section{Analyses}
Since LiST performs label-free test-time search, an important question is whether the optimized energy is aligned with the downstream task metric. We analyze this by comparing the energy change of candidate simplex perturbations with their corresponding metric change, both measured relative to the task-conditioned prior. Specifically, for a candidate perturbation $\delta$, we define
\begin{equation}
\Delta \mathcal{E} = \mathcal{E}(\delta)-\mathcal{E}(0)
\end{equation}
where smaller value indicates larger energy decrease, and measure the corresponding task-metric change $\Delta \mathrm{Metric}$.

Figure~\ref{fig:energy_alignment}(a) bins candidate perturbations by $\Delta \mathcal{E}$. Candidates with larger energy decreases yield higher average metric gains, while energy-increasing candidates lead to negative changes. This shows that the proposed prompt-level energy is an effective surrogate for test-time selection.

However, energy reduction alone does not ensure reliable adaptation. Figure~\ref{fig:energy_alignment}(b) shows a candidate-level scatter plot, with each point representing a simplex perturbation and color indicating the maximum normalized constraint violation,
\begin{equation}
\nu(\delta)=\max\{c_{\mathrm{prior}}, c_{\mathrm{dist}}, c_{\mathrm{cons}}\}
\end{equation}
Although many low-energy candidates improve the task metric, some low-energy candidates with large constraint violations still degrade performance. These harmful candidates usually deviate from the target prior, move outside the local task geometry, or produce unstable stochastic predictions. This observation explains why LiST does not simply deploy the lowest-energy candidate. Instead, the safe acceptance rule accepts a searched candidate only when it achieves sufficient energy improvement while remaining feasible; otherwise, the model falls back to the task-conditioned prior.

Overall, the analysis shows that energy decrease is useful but must be interpreted under feasibility constraints. The prior, geometry, and consistency constraints make the energy signal more reliable for final prediction. More analyses is in Section~\ref{More}.

\section{Conclusion}
In this paper, we introduced LiST, a local-simplex framework for label-free test-time LoRA fusion. LiST converts a frozen adapter bank into a target-conditioned local search space, performs branch-preserving fusion, and optimizes only low-dimensional simplex weights using a prompt-level energy with prior, geometry, and consistency constraints. A safe acceptance rule further prevents unreliable low-energy solutions by falling back to the task-conditioned prior. Experiments on multimodal and language benchmarks show that LiST improves over static merging and conventional TTA baselines while preserving task-specific adapter utility.

\section*{Limitations}
\label{sec:limitation}
LiST assumes access to a bank of related task-specific LoRA adapters, and its benefit may be limited when no useful neighboring adapters are available. 
The test-time search introduces additional inference cost compared with directly using a single adapter, although the search is low-dimensional and all model parameters remain frozen. 
Finally, the prompt-level energy is a surrogate objective and may be less aligned with task metrics for highly open-ended generation tasks.
\section*{Ethical Considerations}
\label{sec:ethics-impact}
LiST operates entirely on publicly available pretrained models and benchmarks, and introduces no new data collection. The method reduces per-task compute by reusing frozen adapters rather than retraining, lowering the carbon footprint of multi-task deployment. We foresee no direct ethical risks specific to LoRA composition; standard concerns around model misuse and bias inherited from base models apply equally here.

\section*{Acknowledgement}
\label{sec:acknowledgement}
This work is supported by Hong Kong RGC General Research Fund (Grant Nos. 15221123, 15216424, and 15211525), and the Hong Kong PolyU Internal Research Fund (Grant Nos. P0058468 and P0062168).

\bibliography{custom}

\appendix

\section{Appendix}
\label{sec:appendix}
\subsection{Benchmark Details}
\label{benchmark}
The GLUE benchmark includes SST-2, MRPC, RTE, QNLI, QQP, CoLA, MNLI-m, MNLI-mm, STS-B, and WNLI. MM-MergeBench includes SciQA, Image, VQA, REC, OCR, VizWiz, Flickr, IconQA, AVQA, Image-R, S2W, and TabMWP. We further divide MM-MergeBench into seen and unseen domains: the seen domains include SciQA, Image, VQA, REC, OCR, VizWiz, Flickr, and IconQA, while the unseen domains include AVQA, Image-R, S2W, and TabMWP. Unseen domains contain only test samples and provide no training samples for training LoRA adapters. Consequently, no target-domain LoRA adapter is trained or used for AVQA, Image-R, S2W, or TabMWP. During unseen-domain evaluation, LiST reuses only the LoRA adapters trained on seen domains and constructs a local simplex from retrieved seen-domain adapters. This setting evaluates cross-domain generalization of adapter composition rather than adaptation with an unseen-domain LoRA.

\subsection{Hyperparameter Sensitivity}
\label{sec:ablation}
\textbf{Sensitivity to Neighbor Count.}
Table \ref{tab:sensitivity_k} analyzes the effect of the number of neighboring adapters $K$. When $K=0$ , the model reduces to the target-task adapter and avoids negative transfer, but it cannot use complementary information from related tasks. A very small neighborhood, such as $K=1$, is not sufficiently expressive and may still select an imperfect neighbor, leading to degraded performance. The best result is obtained at $K=3$, while $K=5$ gives nearly identical performance with slightly higher latency. However, further increasing $K$ hurts both accuracy and negative transfer, and using all adapters leads to the worst performance among fusion settings. This trend supports the local-simplex assumption: useful transfer comes from a small set of related adapters, whereas large or global neighborhoods introduce unrelated task directions and increase computational cost.

\noindent \textbf{Sensitivity to Monte Carlo Forward Count.}
Table \ref{tab:sensitivity_m} studies the number of stochastic forward passes used for estimating the prompt-level energy and stochastic consistency. With only one forward pass, the energy estimate is noisy and the performance is poor. Increasing $M$, from 2 to 4 substantially improves performance and JS estimate stability, showing that multiple stochastic evaluations are important for reliable candidate scoring. The performance continues to improve at $M=8$, where the JS estimate becomes highly stable. Increasing $M$ further to 16 brings only marginal improvement but introduces additional latency. We therefore use $M=8$ as the default setting, which provides a good balance between stability, accuracy, and inference cost.

\noindent \textbf{Sensitivity to Acceptance Margin.}
Table \ref{tab:sensitivity_acceptance_margin} evaluates the energy-improvement margin $m_{accept}$ in the safe acceptance rule. When the margin is zero, the model accepts many searched candidates, but this leads to low performance and high negative transfer, suggesting that small energy reductions are not always reliable. Increasing the margin to 1.0 improves both performance and robustness by filtering weak improvements. The best result is achieved at $m_{accept}=3.0$ where the accept rate remains moderate and negative transfer is substantially reduced. A larger margin, $m_{accept}=5.0$, becomes too conservative: it rejects many beneficial candidates and slightly lowers the final performance. These results show that the acceptance margin controls a trade-off between exploiting useful test-time search results and avoiding unreliable low-energy candidates.

\begin{table}[t]
\centering
\resizebox{0.45\textwidth}{!}{%
\begin{tabular}{l|ccc}
\toprule
\textbf{$K$} & \textbf{Avg.} & \textbf{Neg. Transfer $\downarrow$} & \textbf{Latency} \\
\midrule
0 & 87.35 & \textbf{0.00} & \textbf{123.32} \\
1 & 85.96 & 0.13 & 306.81 \\
3 & \textbf{87.70} & 0.02 & 318.56 \\
5 & 87.68 & 0.04 & 329.62 \\
7 & 84.92 & 0.10 & 340.56 \\
\bottomrule
\end{tabular}}
\vspace{-0.6em}
\caption{\textbf{Sensitivity analysis on neighbor count $K$.}}
\label{tab:sensitivity_k}
\end{table}

\begin{table}[t]
\centering
\resizebox{0.45\textwidth}{!}{%
\begin{tabular}{l|ccc}
\toprule
$M$ & \textbf{Avg} & \textbf{JS Estimate Stability $\uparrow$} & \textbf{Latency} \\
\midrule
1 & 80.92 & -- & \textbf{172.88} \\
2 & 83.19 & 94.28 & 244.23 \\
4 & 86.83 & 96.07 & 292.94 \\
8 & 87.70 & 98.89 & 331.53 \\
16 & \textbf{87.72} & \textbf{99.02} & 378.64 \\
\bottomrule
\end{tabular}}
\vspace{-0.6em}
\caption{\textbf{Sensitivity of MC forward count $M$.}}
\label{tab:sensitivity_m}
\end{table}
\begin{table}[t]
\centering
\resizebox{0.45\textwidth}{!}{%
\begin{tabular}{l|ccc}
\toprule
$m_{\mathrm{accept}}$ & \textbf{Avg} & \textbf{Accept Rate} & \textbf{Neg. Transfer $\downarrow$} \\
\midrule
0.00 & 80.85 & \textbf{57.63} & 0.132 \\
1.00 & 83.69 & 53.37 & 0.073 \\
3.00 & \textbf{87.70} & 42.76 & \textbf{0.024} \\
5.00 & 86.80 & 11.96 & 0.025 \\
\bottomrule
\end{tabular}}
\vspace{-0.6em}
\caption{\textbf{Sensitivity of acceptance margin $m_{\mathrm{accept}}$.}}
\label{tab:sensitivity_acceptance_margin}
\end{table}

\begin{figure}[t]
    \centering
    \includegraphics[width=\linewidth]{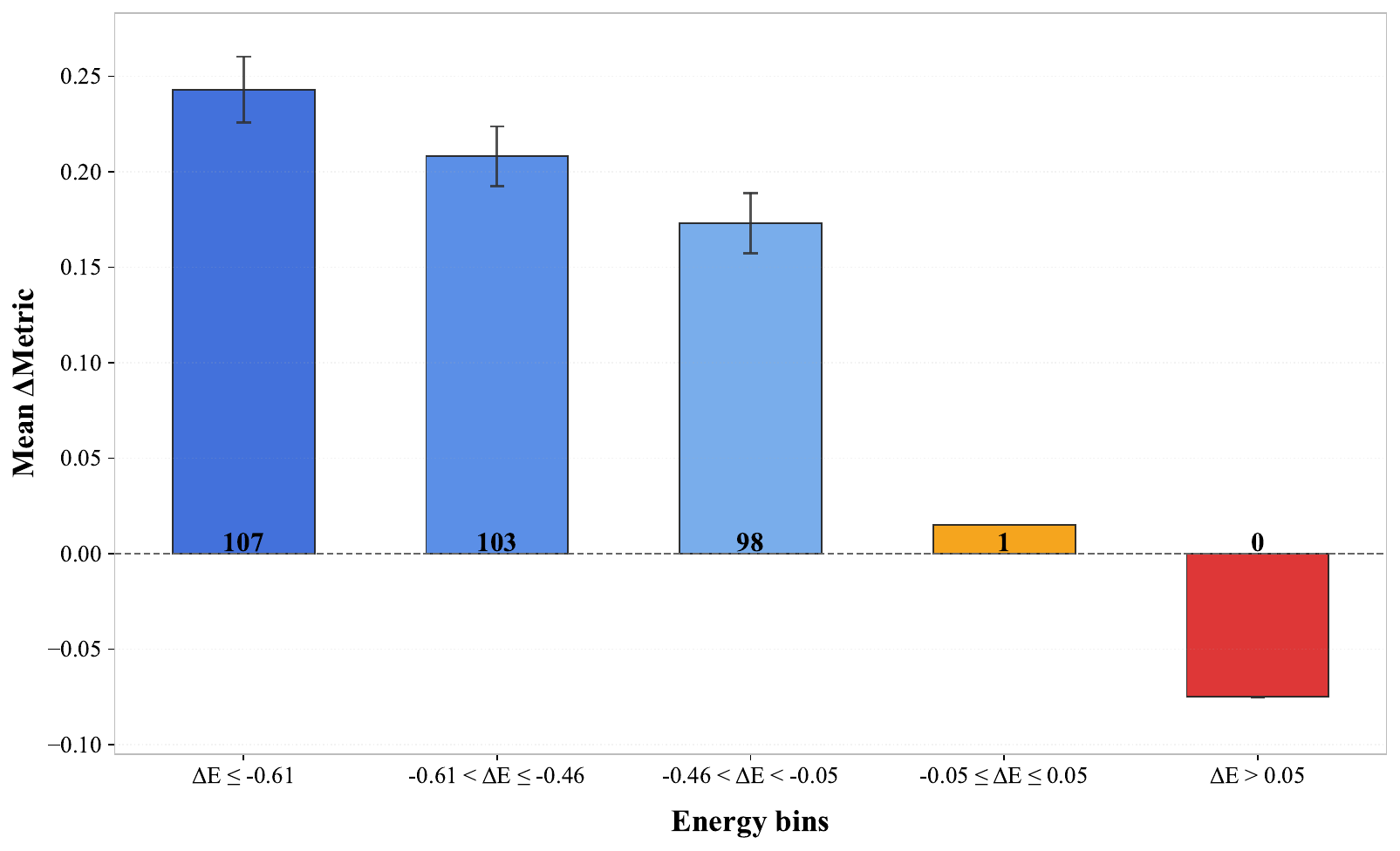}
    \caption{
    Energy--performance alignment on feasible candidates only.
    After filtering high-violation candidates, larger energy decreases correspond more clearly to downstream metric gains, supporting the feasibility-aware interpretation of the prompt-level energy.
    }
    \label{fig:energy_feasible}
\end{figure}
\begin{figure}[t]
    \centering
    \includegraphics[width=\linewidth]{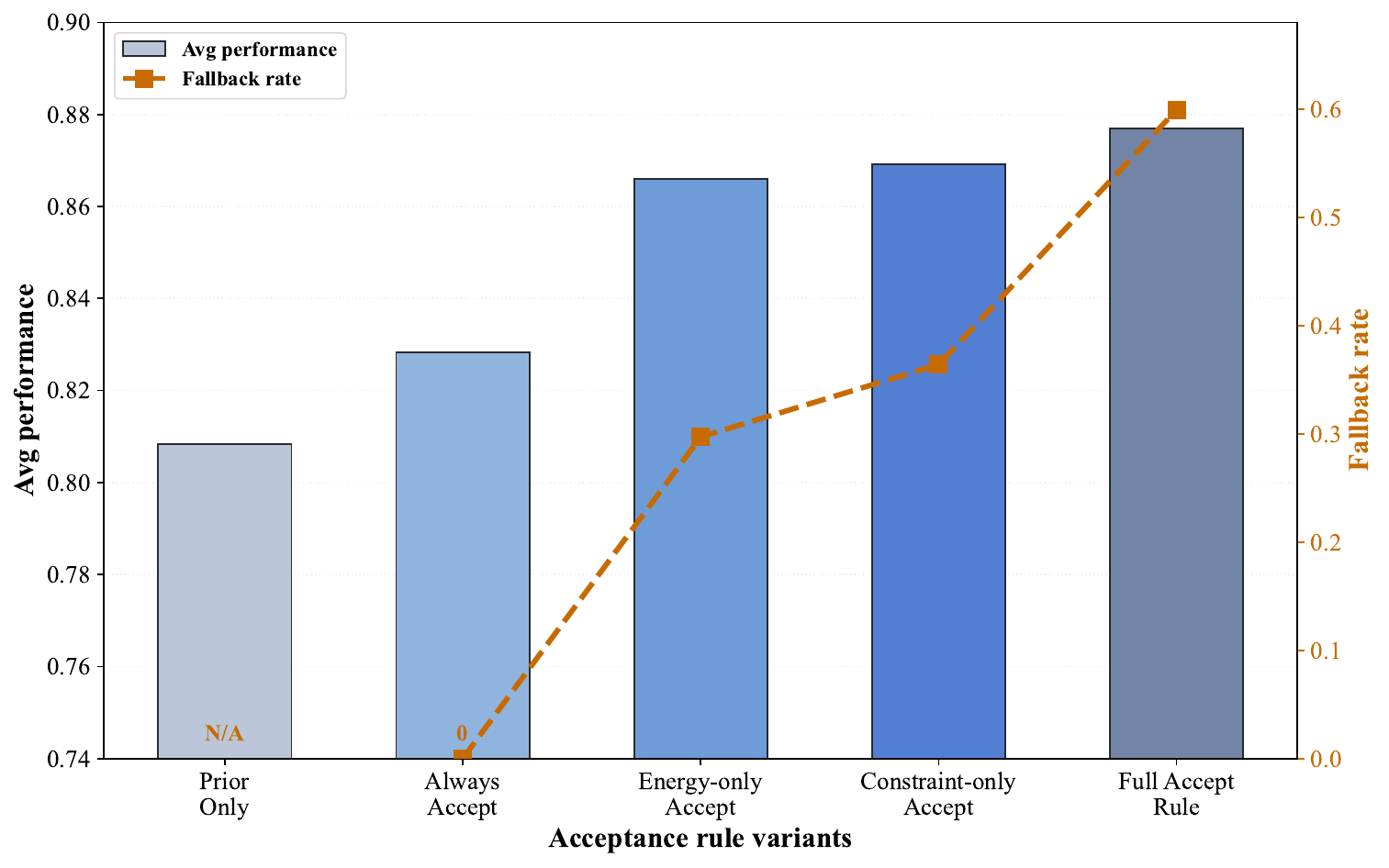}
    \caption{
    Effect of safe acceptance rules.
    The full rule achieves the best average performance while triggering more fallback, showing that feasibility-aware rejection of unreliable searched candidates improves robustness.
    }
    \label{fig:acceptance_rule}
\end{figure}

\section{More Analyses}
\label{More}
\paragraph{Energy--performance alignment on feasible candidates.}
In the main analysis, we show that energy reduction is generally aligned with downstream metric improvement, but this alignment can be weakened by candidates with large constraint violations. We further examine this relationship after applying the feasibility filter used by the safe acceptance rule. Figure~\ref{fig:energy_feasible} groups feasible candidate perturbations by their relative energy change,
$\Delta \mathcal{E}=\mathcal{E}(\delta)-\mathcal{E}(0)$.
After high-violation candidates are removed, larger energy decreases show a cleaner positive association with metric gains. This supports the role of prior, geometric, and stochastic-consistency constraints in making the prompt-level energy a more reliable test-time selection signal.

\paragraph{Effect of safe acceptance rules.}
We also visualize how different deployment rules affect final performance and fallback behavior. Figure~\ref{fig:acceptance_rule} compares prior-only inference, always accepting searched candidates, energy-only acceptance, constraint-only acceptance, and the full acceptance rule. Always accepting searched candidates improves over the prior-only baseline, confirming that test-time search can find useful sample-specific LoRA compositions. However, the full acceptance rule achieves the best average performance while triggering more fallback, indicating that fallback is not merely conservative. Instead, it rejects unreliable searched candidates whose energy improvement is not supported by feasibility constraints.

\section{LLM Usage}
\label{sec:llm_usage}
We used ChatGPT and Claude to assist with drafting and refining text. All AI-assisted content was reviewed, revised, and verified by the authors. These tools contributed to wording and phrasing suggestions only, and played no independent role in research ideation, experimental design, or data analysis. The authors bear full responsibility for all content in this manuscript.
\end{document}